\documentclass[letterpaper]{article} % DO NOT CHANGE THIS
\usepackage{aaai2026}
\usepackage{times}  % DO NOT CHANGE THIS
\usepackage{helvet}  % DO NOT CHANGE THIS
\usepackage{courier}  % DO NOT CHANGE THIS
\usepackage[hyphens]{url}  % DO NOT CHANGE THIS
\usepackage{graphicx} % DO NOT CHANGE THIS
\usepackage{natbib}  % DO NOT CHANGE THIS AND DO NOT ADD ANY OPTIONS TO IT
\usepackage{caption} % DO NOT CHANGE THIS AND DO NOT ADD ANY OPTIONS TO IT
\usepackage{algorithm}
\usepackage{algorithmic}

\usepackage{microtype}
\usepackage{booktabs}
\usepackage{amsmath}
\usepackage{lineno}
\usepackage{graphicx}
\usepackage{multirow}
\usepackage{csquotes}
\usepackage{listings}
\usepackage{colortbl}
\usepackage{pifont}
\usepackage{xcolor} % 支持颜色
\usepackage{fancyvrb} % 允许在 Verbatim 环境中使用颜色命令

\definecolor{darkblue}{rgb}{0, 0, 0.5}

\usepackage{amsfonts}  
\usepackage{array}

\usepackage{newfloat}
\usepackage{listings}
\DeclareCaptionStyle{ruled}{labelfont=normalfont,labelsep=colon,strut=off} % DO NOT CHANGE THIS
\floatstyle{ruled}
\newfloat{listing}{tb}{lst}{}
\floatname{listing}{Listing}
\title{Collaborative LLM Numerical Reasoning with Local Data Protection}

\author{
Min Zhang\textsuperscript{1,2}\thanks{Most work done during Min's internship at AWS AI.}, 
Yuzhe Lu\textsuperscript{1}, 
Yun Zhou\textsuperscript{1}, 
Panpan Xu\textsuperscript{1}, \\
Lin Lee Cheong\textsuperscript{1}, 
Chang-Tien Lu\textsuperscript{2}, 
Haozhu Wang\textsuperscript{1}
}
        
\affiliations {
    \textsuperscript{\rm 1}AWS AI
    \textsuperscript{\rm 2}Virginia Tech
}

\usepackage{bibentry}
\begin{document}

\maketitle

\begin{abstract}

Numerical reasoning over documents, which demands both contextual understanding and logical inference, is challenging for low-capacity local models deployed on computation-constrained devices.
Although such complex reasoning queries could be routed to powerful remote models like GPT-4, exposing local data raises significant data leakage concerns.
Existing mitigation methods generate problem descriptions or examples for remote assistance.
However, the inherent complexity of numerical reasoning hinders the local model from generating logically equivalent queries and accurately inferring answers with remote guidance.
In this paper, we present a model collaboration framework with two key innovations: (1) a context-aware synthesis strategy that shifts the query topics while preserving reasoning patterns; 
and (2) a tool-based answer reconstruction approach that reuses the remote-generated plug-and-play solution with code snippets.
Experimental results demonstrate that our method achieves better reasoning accuracy than solely using local models while providing stronger data protection than fully relying on remote models.
Furthermore, our method improves accuracy by 16.2\% - 43.6\% while reducing data leakage by 2.3\% - 44.6\% compared to existing data protection approaches.

\end{abstract}

% Although some remote black-box models are powerful, directly transmitting the raw local data to the remote black-box model leads to concerns about local data leakage. While some local data protection methods have shown good performance in tasks like text classification and writing, they struggle to effectively address complex logical reasoning over documents that involve localizing information from the original text and performing logical reasoning. It is difficult to strike a balance between model utility and local data protection, which requires hiding sensitive information while keeping sufficient information for model utility.
% It is also hard to generate the high-level description or a similar example to query the remote model due to the difficulty for the local model to understand the key point of the problem. It is also difficult for the local model to recover the original answer from synthesized information due to the limited learning ability to perform the complex logical reasoning. 
% In this paper, we propose a method of topic rewriting and numeric switches to achieve local data anonymization and recovery. Our experiments demonstrate that, compared to existing data protection methods, our approach not only effectively reduces data leakage but also significantly improves reasoning accuracy.

\section{Introduction}

% Model cascades \cite{yuelarge, chen2023frugalgpt} leverage the collaboration between a local model and a more powerful black-box remote model to compensate for the shortcomings of the local model. 
% Typically, more difficult problems and corresponding contexts are sent to the remote model for resolution. 

Numerical reasoning over documents is a practical yet complex task that often requires powerful black-box models like GPT-4 for problem-solving~\cite{akhtar-etal-2023-exploring-nr}. This task demands a deep understanding of documents, the ability to identify relationships from scattered evidence, and the capability to derive answers through quantitative calculations. 
In real-life scenarios, numerical reasoning is essential in tasks like analyzing financial reports~\cite{chen2021finqa,zhao2022multihiertt,ma2025llm_nr}, research papers~\cite{wu2025agentic_nr}, medical documents~\cite{mahendra-etal-2024-numbers_nr}, and contracts with numerical conditions~\cite{huangnumerical_nr}.
Due to the demanding requirements for contextual understanding and logical reasoning, on-device or in-house small models often struggle to solve such problems effectively. Consequently, remote black-box models with strong problem-solving capabilities are frequently accessed via API calls to address these challenges.

\begin{figure*}[t]
    \begin{center}
    \includegraphics[width=6.6in]{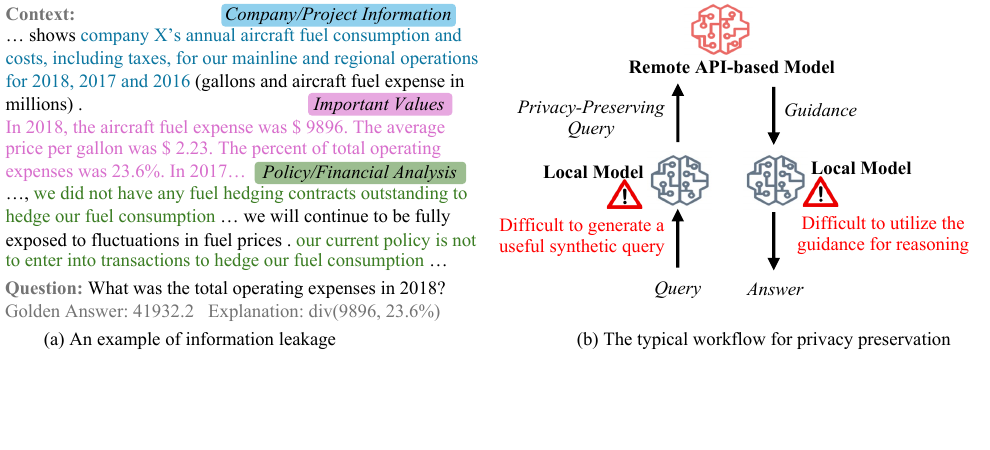}
    \vspace{-19mm}
    \end{center}
    \caption{(a) An example from FinQA with highlighted sensitive information. (b) The typical workflow and challenges for preserving privacy in interactions with remote API-based models.}
    \label{fig: example}
\end{figure*}

However, the direct exposure of local data to remote models introduces significant risks of information leakage \cite{wang2024pandora}. 
Following prior studies \cite{zhou2023textobfuscator, tong2023inferdpt, hartmann2024can}, we define local privacy information to pertain to every word, excluding non-sensitive stop words \cite{yue2021differential}.
Sensitive data can be presented explicitly or embedded implicitly within various contexts and formats, including company details, operational values, and strategic analyses, as illustrated in Fig.~\ref{fig: example} (a).
While some works \cite{siyan2024papillon, chen2023hide, Aahill2023AzureAI} detect and remove explicit Personally Identifiable Information (PII), models like GPT-4 can infer personal attributes from residual context \cite{staabbeyond}.
Similarly, the sentence \textit{\enquote{our current policy is not to enter into transactions to hedge our fuel consumption...}} in the example reveals confidential policy decisions without containing any explicit PII. 
Therefore, in this work, we aim to thoroughly protect the document-level local text before the black-box model inference.

The inherent complexity of numerical reasoning over documents exacerbates the trade-off between data leakage and model utility. 
On one hand, local data must be concealed from the remote model to minimize information leakage. On the other hand, the remote model often requires detailed contextual information to have a deep understanding of the problem and to provide accurate help for effective reasoning.
% While some methods aiming to protect privacy in the black-box inference scenario have shown effectiveness in simpler tasks such as classification, summarization, and writing, they fall short in addressing the demands of complex reasoning. 
We identify two primary challenges for reasoning in the typical data protection workflow as shown in Fig.~\ref{fig: example} (b):

\textbf{Difficulty in generating logically coherent synthesized queries.} 
Existing methods often locally synthesize various queries for data protection, such as high-level descriptions \cite{zhang2024cogenesis}, analogous examples \cite{hartmann2024can}, dp-based permutation or paraphrased documents \cite{utpala2023locally}.
However, the complex logical reasoning within documents hampers the synthesis of accurate descriptions or logical-coherent contexts while keeping local data secret.

\textbf{Local answer reconstruction with limited reasoning ability.}
Existing methods leverage a local model to integrate local data and the remote model's response to generate final response. Although they are effective for semantic-focused tasks such as creative writing, translation and summarization \cite{tong2023inferdpt, zhang2024cogenesis, hartmann2024can}, their performance drops dramatically in reasoning tasks which demands both deep contextual understanding and complex logic. Even with accurate hints or examples, small models often struggle to reconstruct correct answers due to their limited reasoning abilities.

To address these challenges, we propose a novel method that transforms reasoning queries into a different domain while preserving the linguistic and logical patterns.
As illustrated in (Fig. \ref{fig: workflow}), by translating reasoning requests from one domain (e.g., \textit{aircraft fuel consumption}) to another (e.g., \textit{advertising revenue}) while preserving their inferential skeleton (e.g., \textit{deriving the total amount from the individual count and percentage}), we enable secure delegation of reasoning to remote models without compromising sensitive information. This pattern-preserving transformation not only protects privacy but also retains the epistemic scaffolding necessary for solution generalization.

Building on this abstraction, we further introduce a tool-based answer reconstruction strategy where the remote model returns a reusable problem-solving tool with executable code snippets. This plug-and-play construct enables precise local answer recovery through direct numerical substitution, without local model dependence.

Our main contributions are summarized as follows:
\begin{itemize} 
    \item We design a topic-shifted and pattern-preserving data synthesis approach that replaces the semantic surface while keeping the reasoning structure intact as a whole, enabling privacy-preserving delegation to remote models. 
    \item We propose a plug-and-play answer reconstruction method that leverages tool-centric reusable solution paradigm from the remote model, facilitating precise answer recovery via numerical substitution.
    \item We achieve a superior accuracy-privacy trade-off than existing collaborative inference methods with data protection measures, improving accuracy by 16.2\% - 43.6\% while reducing data leakage by 2.3\% - 44.6\%. 
\end{itemize}

\section{Related Work}

% \textbf{Model Cascade.} 
% Due to the limited capabilities of locally deployable models and the strong performance of proprietary API-based models, an increasing number of studies have focused on model collaboration ~\cite{miao2023towards,chen2023accelerating,leviathan2023fast}.
% The model cascade framework firstly resolves the problem by the local model, if the local response is not reliable, the local problem will be routed to the powerful remote model to resolve. 
% ~\cite{chen2023frugalgpt} trains a router model to make the decision.
% \cite{yuelarge} uses the voting score of the most common answer as the criterion. 
% The framework not only reduces the cost of API calls for remote model but also keeps some problems resolved locally which protects the local data from data leakage.
% However, the direct exposure of local data raises concerns about local data leakage. 
% In this paper, we migrate the privacy leakage in the interaction with a remote black-box model.

Prior works address data protection for training data \cite{papernot2017semi, yue2021differential, tian2022seqpate, kurakin2023harnessing, xie2024differentially, yu2024privacy} or for demonstrations \cite{hongdp, carey2024dp}, but these approaches do not protect user queries during inference.

To protect privacy at inference time, some methods add differential privacy (DP) noise to text embeddings in white-box settings \cite{du2023dp, zhou2023textobfuscator}. However, these techniques are incompatible with black-box models, which typically require plain-text inputs rather than embeddings.

% Other studies aim to preserve privacy by detecting and replacing personally identifiable information (PII) such as names, ages, and locations by training a hider model to obfuscate entities with random words \cite{chen2023hide} or optimizing prompts to conceal sensitive attributes \cite{siyan2024papillon}. However, PII replacement is often insufficient, as remaining contextual cues can still leak information \cite{staabbeyond}.

Some studies preserve privacy by replacing PII like names or locations using hider models \cite{chen2023hide} or prompt optimization \cite{siyan2024papillon}, but residual context often still leaks information \cite{staabbeyond}.

For document-level privacy protection, some approaches apply DP-based perturbations to generate semantically similar but altered contexts \cite{tong2023inferdpt, xie2024differentially}, or prompt LLMs to paraphrase documents for downstream tasks like sentiment classification \cite{utpala2023locally}. Yet, these methods generally preserve semantics without guaranteeing logical consistency with the original content. Other works prompt local models to generate high-level descriptions \cite{zhang2024cogenesis} or analogous examples \cite{hartmann2024can}. Still, generating queries that are both privacy-preserving and informative remains challenging, and local models often struggle to effectively utilize remote guidance due to limited capacity.

In contrast, our method enhances both privacy and logical fidelity by introducing a pattern-preserving topic shifter, and further employs a plug-and-play answer reconstruction strategy that avoiding local model dependence.

\section{Method}

\begin{figure*}
    \centering
    \includegraphics[width=6.6in]{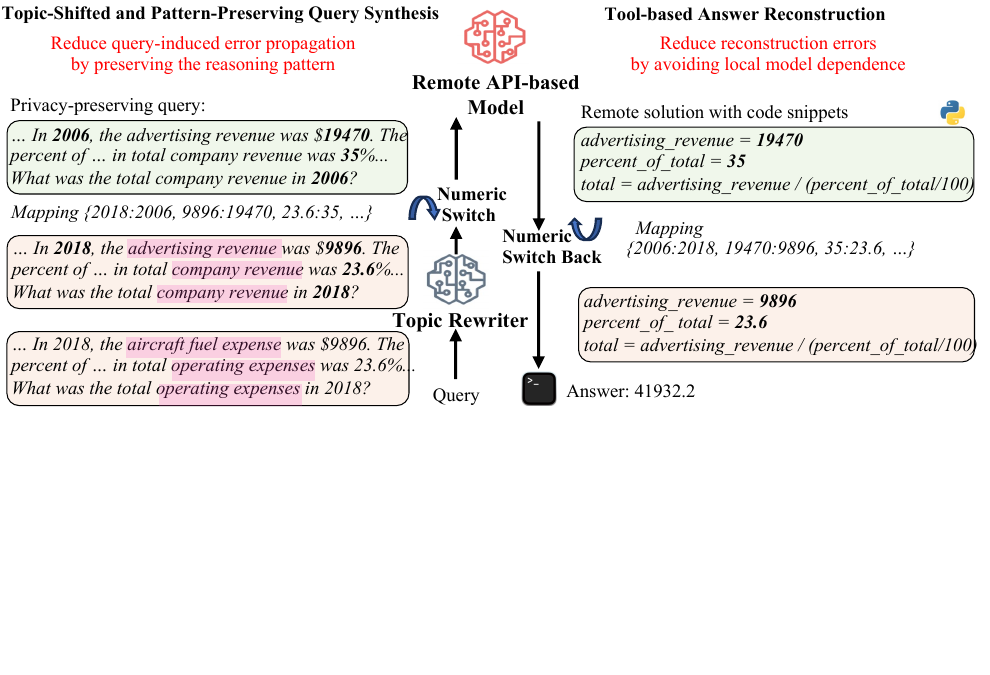}
    \vspace{-43mm}
    \caption{The proposed method llustrated with examples for each step. The original query is transformed from one topic (e.g., \textit{aircraft fuel consumption}) to another (e.g., \textit{advertising revenue}) by a distilled topic rewriter, while preserving its reasoning pattern (e.g., \textit{deriving the total amount from the individual count and percentage}). This enables secure delegation to a remote API-based model to elicit a tool-centric, plug-and-play solution for local answer reconstruction through direct numerical switch, without requiring local model re-inference.}
    % \vspace{-5mm}
    \label{fig: workflow}
\end{figure*}

To maximize the utility of remote models while minimizing local data leakage in numerical reasoning tasks, we propose an effective collaboration protocol based on the observation that sensitive queries can be transformed into a different domain without their underlying mathematical structure altered. 
Our approach involves translating queries from one domain to another, ensuring that the remote model processes semantically transformed but structurally equivalent data. This strategy mitigates the risk of exposing sensitive information while offering the local model a reusable solution.
To achieve this, we introduce two key components of our protocol (as shown in Fig.~\ref{fig: workflow}), topic-shifted and pattern-preserving query synthesis and tool-based answer reconstruction, in the following subsections.

% \subsection{Logic-preserving request synthesis} 
\subsection{Topic-Shifted and Pattern-Preserving Query Synthesis}

In this section, we introduce the module leveraging specialized local models to synthesize requests with shifted topics but equivalent mathematical abstractions. 
Since hiding the local information and maintaining the underlying logic remains challenging for small local models, we fine-tune a dedicated request synthesis model and subsequently apply a numerical replacement strategy.
By decoupling semantic protection and numerical protection, 
maintaining numerical values intact in the topic shifter not only facilitates mathematical abstraction consistency verification but also paves the way for local answer reconstruction (as explained in the next section).
Our data synthesis approach is detailed below.

\subsubsection{Topic Shifter}
To protect the overall local information instead of detecting specific sensitive words, we instruct the synthesis model to shift the topic while maintaining the original format, logic and numerical values. 
By preserving numeric values, we establish a clear mapping between the original and synthesized objects (e.g., \textit{fuel expense} $\leftrightarrow$ \textit{advertising revenue} in Fig.~\ref{fig: workflow}).
% The corresponding prompt and example are provided in Appendix~\ref{sec:appendix-topic-shifter}.
We formally characterize the input and output of the proposed request synthesis model, denoted as $\mathcal{M_S}$, using the following equation:
\begin{equation}
    \quad \tilde{C}, \tilde{q} = \mathcal{M_S}(C, q)
\end{equation}
where $\tilde{C}, \tilde{q}$ represent the transformed context and query derived from the original inputs $C, q$.

% For numerical reasoning over long documents, we note that $C = \{s_1, s_2, \dots, s_n\}$ is only a subset of the full document, where $s_i$ are retrieved sentences relevant to the user query $q$. 
% This shortened context is a natural choice as input to the synthesis model because the key sentences can be extracted alongside the local inference, and significantly simplifies the transformation compared to processing the entire document. 
% Details of the local retriever can be found in the Appendix~\ref{sec:appendix-local_retriever}, and it can also be replaced with other retrievers.

To deploy an efficient local synthesizer, we distill the capabilities of a large remote model into a smaller local model. The remote model $\mathcal{M_R}$ generates synthetic data for training.
Since the numerical values remain unchanged during the topic shift, a logically consistent rewritten question should yield the same answer as the original question.
Through such post-hoc analysis, we confirm its proficiency in instruction-following and generating pattern-preserving requests. 
In contrast, small local models often struggle to meet these requirements, especially in keeping numerical consistency and logical coherence. Via instruction tuning, we enhance the local synthesizer's rewriting and instruction-following capabilities.

\subsubsection{Data Switch} 
Since the topic shifter focuses on semantic protection, leaving the numerical values unchanged, we further obfuscate numerical values in the synthesized request to ensure complete anonymization.
Using regular expressions, we extract all numbers $\mathcal{N} = \{n_1, n_2, \dots, n_k\}$ from the request and apply a transformation 
\begin{equation}
    h: n_i \mapsto \tilde{n_i}
\end{equation}
We employ three strategies to ensure data transformation quality: special number handling, offset transformation for year-related values, and order-preserving transformation. 
Special numbers (e.g., 28–31 for month-end dates) remain unchanged to preserve their semantic meaning. Integers between the year-related range undergo offset-based transformation that maintains relative differences between years. All other numbers are sorted into intervals and mapped to randomly sampled values from a target range while preserving their original order relationships.
% Details can be found in Appendix.
The final synthesized request with transformed numerical values, denoted as $(\tilde{C_h}, \tilde{q_h})$, is then forwarded to the remote model for assistance.

The data switch ensures numerical security, complementing the topic shifter’s for comprehensive data protection. Decoupling numerical protection during local data synthesis, we simplify subsequent local reconstruction. Serving as a bridge between the topic shifter and local reconstruction, the data switch ensures a seamless transition for accurate and secure data processing.

By fine-tuning the specialized local synthesizer $\mathcal{M_S}$, we ensure the synthetic request retains the original problem-solving logic while achieving a complete topic shift. Additionally, the numeric transformation step guarantees that shared values do not expose sensitive information. The transformation $h$ is stored locally as a dictionary, and its role in enhancing local inference accuracy is further elaborated in the subsequent section.

\subsection{Tool-based Local Answer Reconstruction}

In addition to protecting local data, another key aspect of privacy-preserving collaborative inference is how to best leverage remote assistance. Due to privacy constraints, the solution from the remote model to a proxy request cannot be used directly. Naively, one could simply add the remote guidance to local model's context to elicit a better response. However, we found that this strategy does not lead to satisfactory performance for the local model on numerical reasoning tasks owing to its limited capabilities. Thus, we propose a more structured scheme for the local model to generate its answers, which leads to dramatic performance improvements.

% Existing method prompts the remote model to generate a solution for the synthesized example or some high-level hints as the guidance information. Then the local model is prompted to conduct local inference with the remote guidance. However, due to the local model's limited reasoning ability, it is hard to generate accurate inferences.

Since the synthesized request maintains the same logic as the original one, they share the same problem-solving pattern. To best preserve and communicate this pattern, we instruct the remote model to generate Python code \cite{chen2022program} as a tool for the local model. During answer reconstruction, the local model will simply perform data substitution: since intermediate steps are represented using Python variables, substituting the input data is sufficient. The final answer is obtained by executing the code in an interpreter. We elaborate each of these steps below.

\subsubsection{Remote Assistance}
After performing topic rewriting and numerical anonymization, we provide the synthesized context $\tilde{C_{h}}$ and synthesized question $\tilde{q_{h}}$ to the remote model $\mathcal{M_R}$:
% \begin{equation}
%     F, f(m_{1}, m_{2} \dots, m_{t}) = \mathcal{M_R}(\tilde{C_{h}}, \tilde{q_{h}}), \qquad \text{where } m_i \in \{\tilde{n_1}, \tilde{n_2}, \dots, \tilde{n_k}\}
% \end{equation}

\begin{equation}
    f(m_{1}, m_{2} \dots, m_{t}) = \mathcal{M_R}(\tilde{C_{h}}, \tilde{q_{h}})
\end{equation}

% \begin{equation}
%      \qquad \text{where } m_i \in \{\tilde{n_1}, \tilde{n_2}, \dots, \tilde{n_k}\}
% \end{equation}

where \( m_i \in \{\tilde{n_1}, \tilde{n_2}, \dots, \tilde{n_k} \} \). The remote model returns the Python code snippets $f(\cdot)$  for problem solving with transformed numeric inputs from the synthesized request.
% where $\text{Code}^{\text{patt}}$ is python code which reflects the problem-solving pattern generated by the remote LLM. 

\subsubsection{Local Reconstruction} Since the local model often fails to recognize the logical connection between the original and synthesized request and thus struggles to write Python codes for the local problem, we implement a plug-and-play approach to reuse the Python solution $f$ from the remote model. Specifically, we compute the answer to the local problem by directly executing the following with Python interpreter: 

\begin{equation}
    f(h^{-1}(m_1), h^{-1}(m_2), \dots, h^{-1}(m_t))
\end{equation}

Recall that $h$ is the mapping between original and transformed numeric values. Simply by swapping the input values, we can obtain the answer for the local problem thanks to the logical consistency between original and synthesized requests. With the collaboration scheme above, we maximize the remote model utility while relieving the reasoning model from reasoning burdens. Before we move on to showcase the superior performance of our method, we would like to emphasize the unique contribution of our approach. While our method might seem to be an instantiation of Program-of-Thought (PoT)\cite{chen2022program}, our focus is mainly on how to reuse the Python solution with quite different contexts and accurately transfer to a logically equivalent problem instead of leveraging code to enhance reasoning.
It's important to note that PoT is merely a vehicle for our problem-solving pattern, and its success is deeply rooted in the logical consistency of queries as well as the decoupling of semantics and data protection within our approach.

% \textbf{Local Reconstruction} We then use the numerical dictionary to reverse substitute the numerical values in the anonymized Python solution, recovering the original answer to the question. 
% \begin{equation}
%     \text{Code}_f
%      = \text{Number Switch Back}(\text{Code}^{\text{patt}}, V_{\text{map}})
% \end{equation}
% where $\text{Code}_f$ is the Python code for the local question. Then the local answer can be obtained by executing the code in Python interpreter:
% \begin{equation}
%     \quad
%      A_f = \text{Exec}(\text{Code}_f)
% \end{equation}
% where $A_f$ is the execution result.
% This approach ensures accuracy meanwhile avoiding local reasoning and additional computation, given the limited learning capacity of the small model.

% 我们方法的创新是，通过创建逻辑一致的问题，我们可以从remote model获取可复用的问题解决模式，然后通过数据替换来直接复用这个问题解决模式，从来达到精确的本地答案重建。
% While our method utilizes the concept of Program-of-Thought (PoT) \cite{chen2022program} to generate Python code as a reasoning solution, it introduces significant innovations that distinguish it from prior work. Unlike conventional PoT approaches, which primarily focus on generating code for direct problem-solving, our method leverages synthesized questions that maintain the same logical structure as the original queries. Our innovation lies in creating logically consistent questions to obtain reusable problem-solving patterns from the remote model, which are then directly reused through data substitution to achieve accurate local answer reconstruction.

\section{Experiment Settings}

\begin{table*}[]
\centering
\begin{tabular}{cccccc}
\toprule
\multicolumn{2}{c|}{Datasets}                                                                                                                 & \multicolumn{2}{c|}{MultiHiertt}                                       & \multicolumn{2}{c}{FinQA}                         \\ \hline
\multicolumn{2}{c|}{Metric}                                                                                                                 & \multicolumn{1}{c|}{Acc(\%)$\uparrow$}        & \multicolumn{1}{c|}{Leakage(\%)$\downarrow$}         & \multicolumn{1}{c|}{Acc(\%)$\uparrow$}        & Leakage(\%)$\downarrow$         \\ \toprule
\multicolumn{6}{c}{Local Model: Phi-3-mini-128k-instruct (3.8B)}                                                                                                                                                                                                           \\ \hline
\multicolumn{1}{c|}{\multirow{2}{*}{\begin{tabular}[c]{@{}c@{}}Local\\ Methods\end{tabular}}}         & \multicolumn{1}{c|}{Single-inference} & \multicolumn{1}{c|}{54.4}          & \multicolumn{1}{c|}{0}            & \multicolumn{1}{c|}{71.1}          & 0            \\ \cline{2-6} 
\multicolumn{1}{c|}{}                                                                                 & \multicolumn{1}{c|}{Self-consistency} & \multicolumn{1}{c|}{64.6}          & \multicolumn{1}{c|}{0}            & \multicolumn{1}{c|}{80.0}          & 0            \\ \hline
\multicolumn{1}{c|}{\multirow{3}{*}{\begin{tabular}[c]{@{}c@{}}Collaboration\\ Methods\end{tabular}}} & \multicolumn{1}{c|}{Hint}             & \multicolumn{1}{c|}{42.7}          & \multicolumn{1}{c|}{6.0}          & \multicolumn{1}{c|}{63.9}          & 28.8         \\ \cline{2-6} 
\multicolumn{1}{c|}{}                                                                                 & \multicolumn{1}{c|}{Example}          & \multicolumn{1}{c|}{57.0}          & \multicolumn{1}{c|}{16.1}         & \multicolumn{1}{c|}{71.4}          & 38.9         \\ \cline{2-6} 
\multicolumn{1}{c|}{}                                                                                 & \multicolumn{1}{c|}{Ours}             & \multicolumn{1}{c|}{\textbf{80.1}} & \multicolumn{1}{c|}{\textbf{3.7}} & \multicolumn{1}{c|}{\textbf{87.6}} & \textbf{6.4} \\ \bottomrule
\multicolumn{6}{c}{Local Model: Llama-3.2-3B-Instruct}                                                                                                                                                                                                                     \\ \hline
\multicolumn{1}{c|}{\multirow{2}{*}{\begin{tabular}[c]{@{}c@{}}Local\\ Methods\end{tabular}}}         & \multicolumn{1}{c|}{Single-inference} & \multicolumn{1}{c|}{33.8}          & \multicolumn{1}{c|}{0}            & \multicolumn{1}{c|}{54.0}          & 0            \\ \cline{2-6} 
\multicolumn{1}{c|}{}                                                                                 & \multicolumn{1}{c|}{Self-consistency} & \multicolumn{1}{c|}{44.5}          & \multicolumn{1}{c|}{0}            & \multicolumn{1}{c|}{68.6}          & 0            \\ \hline
\multicolumn{1}{c|}{\multirow{3}{*}{\begin{tabular}[c]{@{}c@{}}Collaboration\\ Methods\end{tabular}}} & \multicolumn{1}{c|}{Hint}             & \multicolumn{1}{c|}{27.0}          & \multicolumn{1}{c|}{20.9}          & \multicolumn{1}{c|}{55.3}          & 50.5          \\ \cline{2-6} 
\multicolumn{1}{c|}{}                                                                                 & \multicolumn{1}{c|}{Example}          & \multicolumn{1}{c|}{31.1}          & \multicolumn{1}{c|}{18.8}          & \multicolumn{1}{c|}{55.5}          & 20.0          \\ \cline{2-6} 
\multicolumn{1}{c|}{}                                                                                 & \multicolumn{1}{c|}{Ours}             & \multicolumn{1}{c|}{\textbf{70.6}} & \multicolumn{1}{c|}{\textbf{5.5}} & \multicolumn{1}{c|}{\textbf{90.1}} & \textbf{5.9} \\ \bottomrule
\end{tabular}
\caption{Our method outperforms others by improving accuracy with less local data leakage. The accuracy score is normalized against that of the remote model with full local data.}
\label{table: result_multi}
\end{table*}

\subsubsection{Datasets}  
We conduct experiments using two question-answering datasets: FinQA \cite{chen2021finqa} and MultiHiertt \cite{zhao2022multihiertt}. Both datasets involve questions that require numerical reasoning based on provided documents. They contain both explicit and implicit sensitive information, including financial data analysis, project details, and decision-making content, making them well-suited for evaluating our local data protection approach. 
% Further details are provided in Appendix~\ref{sec:appendix-datasets}.  
Further details are provided in Appendix. 

\subsubsection{Data Processing}  
For privacy-preserving collaboration methods, we shorten the context by retrieving relevant parts of the full document.
This retrieval process is a natural choice for long-document processing and reduces the data synthesis burden compared to using the entire document.
To build it, we leverage the local self-consistency inference process by prompting the local model to surface supporting sentences during its reasoning.
These are then combined with results from a BM25 retriever to form the final context.
Details of the local retriever can be found in the Appendix, and it can also be replaced with other retrievers.

\subsubsection{Models}  
We employed two lightweight local models designed for resource-constrained environments. Specifically, we used Phi-3-mini-128k-instruct (3.8B params), and Llama-3.2-3B-Instruct. For collaborative reasoning, we used GPT-4o as the remote model.  

\subsubsection{Model Distillation Settings}  
For the topic rewriter model, we adopted a distillation setup where Llama-3.2-3B-Instruct served as the student model and GPT-4o as the teacher model. After applying a data quality filtering process, via leakage evaluation, conflict evidence detection, and answer consistency verification (see Appendix), we retained 5762 training samples out of 6360 samples from MultiHiertt's training set. We only trained a single synthesis model and used it for different datasets and local models.

\subsubsection{Evaluation Metrics} 
Our evaluation comprises two key aspects: accuracy and local data leakage. Accuracy assesses whether the predicted answer matches the ground truth. Following \cite{hartmann2024can}, we report normalized accuracy scores (actual accuracy divided by remote-only inference accuracy) to show how much of the remote model's performance different methods can achieve. For data leakage assessment, we follow prior studies \cite{zhou2023textobfuscator, tong2023inferdpt}, which define local privacy information as pertaining to every word, excluding non-sensitive stop words \cite{yue2021differential}. 
Similar to recent LLM-as-a-judge approaches \cite{hartmann2024can, siyan2024papillon} for sensitive data leakage detection, we prompt the LLM to provide a binary judgment on whether the synthesized data contains original information.  
Specifically, we instruct a strong model to evaluate the presence of local information in remote model interactions. The evaluation prompt is as follows: \textit{Given context A and context B, determine whether context B uses information from context A. Ignore table formats and sentence structures; If they share some similar important nouns, it can be considered that context B uses information from context A. Respond directly with Yes or No.} Here, context A denotes the original input, while context B represents the transmitted text during interactions with the remote model.  
We utilized GPT-4o-mini as the judge for data leakage evaluation. To validate the evaluation's reliability, we conducted a manual review of 200 randomly selected leakage mapping results from the training set in both datasets. The agreement rate with human annotations was 96\%, demonstrating a high alignment between the automated evaluation and human judgment.

\subsubsection{Baselines}  
We compare our method with baselines involving local-only approaches, vanilla cascading without data protection, and model collaboration with different protection strategies. All methods utilize 3-shot prompting. We employ top-$p$ sampling \cite{holtzmancurious} with $p=0.9$ for local models and greedy sampling for the remote model. 

\textit{Single Inference:} We use in-context learning with few-shot demonstrations \cite{brown2020language} and Program-of-Thought (PoT) \cite{chen2022program} for local solution generation. Results are averaged over seven runs.

\textit{Self-Consistency:} To enhance local inference, we adopt self-consistency \cite{wang2022self}, selecting the answer with the highest voting consistency from seven executions.

\textit{Vanilla Model Cascading:} Following \cite{yuelarge}, if local answer consistency falls below a threshold, the request is directly sent to a remote black-box model without data protection.

\textit{Hint:} Following \cite{zhang2024cogenesis, hartmann2024can}, the local model generates a problem description, and the remote model provides a high-level hint. The hint is then integrated with local data for local inference.

\textit{Example:} Following \citet{hartmann2024can, utpala2023locally}, we use a local model to rephrase queries by concealing sensitive information (including both topics and numerical values) before sending to a remote model. The local model then combines local data with the remote solution for final inference.

% \yun{our main result seems a lot better than all baselines, may benefit by adding one sentence here on "why we think the baseline selection is comprehensive already" and proactively state the comparison's hiccup if there is any, to avoid reviewer question}

\section{Results}
% In Section \ref{sec:no_referral}, we first show the accuracy-privacy trade-off. Later in Section \ref{sec:referral}, we report the accuracy-privacy trade-off with an active referral module to reflect more realistic settings of model cascades. Then, in Section \ref{sec:ablation}, we provide a thorough ablation study to highlight the importance of the proposed modules in our framework.

We first present the main results, comparing our method with local-only and existing privacy-preserving baselines in terms of accuracy and leakage relative to directly sending queries to the remote model.
We then evaluate different privacy strategies under a model cascade framework, where a decision maker routes only harder queries to the remote model, to assess effectiveness on queries of different difficulty.
Next, we conduct an ablation study to assess the contribution of each component.
We further examine the impact of training data size on model performance.
Finally, we perform an error analysis on incorrect cases.

\subsection{Main Results} 
\label{sec:no_referral}

Table~\ref{table: result_multi} shows the accuracy and leakage results for various methods with different local models on two datasets. Compared to local inference methods, our method provides an alternative solution that trades minimal data leakage for substantial accuracy improvements. 
Compared to collaborative inference methods with other data protection strategies, our method presents a superior improvement in both accuracy and data protection. 

% Compared to collaborative inference methods, our method presents a superior trade-off than model-cascades with data protection measures. 

% \textbf{Bridging local and remote accuracy with minimal data leakage.} Our method effectively enhances local model accuracy to approach remote accuracy with minimal data exposure.

\textbf{Our method significantly improves local accuracy while approaching the upper accuracy bound of fully utilizing the remote LLM, with minimal data leakage.}
% We do not consider the decision-making process in the vanilla model cascade regarding which data should be sent to the remote model. Instead, we forward all pieces of data to the remote model in the collaboration method. The accuracy is how much the method can recover the remote model's ability. On a high level, our method effectively enhances local model accuracy to approach remote performance while ensuring minimal data exposure.
% \textbf{Bridging local and remote accuracy with minimal data leakage.}
% Our method effectively enhances local model accuracy to approach remote performance while ensuring minimal data exposure.
\textit{(1) Improved local accuracy approaching remote accuracy}. Compared to single-inference and self-consistency approaches, our method achieves a notable accuracy boost. For instance, on MultiHiertt, Phi-3-mini-128k-instruct improves from 54.4\% to 80.1\%, and on FinQA, Llama-3.2-3B-Instruct increases from 54.0\% to 90.1\%, closely approaching remote model performance. Our method effectively leverages the remote model to compensate for the limited reasoning capabilities of the local model.
% \textit{Improved local accuracy}. Compared to single-inference and self-consistency approaches, our method achieves a notable accuracy boost. For instance, on MultiHiertt, Phi-3-mini-128k-instruct improves from 54.4\% to 80.1\%, and on FinQA, Llama-3.2-3B-Instruct increases from 54.0\% to 90.1\%. Our method effectively leverages the remote model to compensate for the limited reasoning capabilities of the local model.
% \textit{Approaching remote accuracy}.
% Our method closely approaches remote model performance, demonstrating that effective collaboration can bridge the gap between local and remote capabilities. Particularly, FinQA shows a significant improvement with our method achieving up to 90.1\%, indicating robust performance with both local models.
\textit{(2) Minimal data leakage}.
In addition to substantial accuracy improvements, our method maintains low data leakage. On MultiHiertt, leakage is limited to 3.7\% and 5.5\% for Phi-3-mini-128k-instruct and Llama-3.2-3B-Instruct, respectively. On FinQA, leakage remains at 6.4\% and 5.9\%, reflecting a balanced trade-off between privacy and accuracy. These values are markedly lower compared to complete leakage in remote-only approaches.

% \textbf{New frontier compared to existing data protection methods.}
% Our collaboration method not only improves accuracy but also significantly reduces leakage. 

\textbf{Our method achieves a new frontier in both the accuracy and the data protection over data protection baselines.}
% compared to existing protection measures. This expanded Pareto frontier enables model cascades to support new use cases that were previously infeasible due the stringent privacy-accuracy trade-off.
% This dual advantage makes our method a promising solution for tasks requiring both high accuracy and minimized risk of leakage, demonstrating its effectiveness across different models and datasets.
\textit{(1) Accuracy improvement.} 
Our method demonstrates a substantial performance improvement over existing approaches, with notable gaps observed across various datasets and models. 
Our method outperforms the Hint-based method by 23.7\% - 43.6\% in accuracy, and the Example-based method by 16.2\% - 39.5\%.
Notably, the Hint and Example methods result in even lower accuracy than solely using the local model. This is because the unreliable, logic-incoherent query fails to trigger the remote large model for tailored assistance, and it does not effectively invoke the weaker local model's reasoning capabilities.
In contrast, our method generates logically consistent queries to elicit the remote model to produce problem-solving patterns, and it easily recovers answers through data replacement. 
% These results clearly highlight the effectiveness of our method in delivering superior performance across multiple datasets and models.
\textit{(2) Leakage reduction.}
Our method achieves a significant reduction in data leakage compared to existing approaches. On the MultiHiertt dataset, leakage decreases by 2.3\% - 15.4\%. The reduction is even more pronounced on the FinQA dataset (14.1\% - 44.6\%). 
This is because the Hint method tends to leak contextual information when generating problem descriptions, and the Example method struggles to hide information effectively on its own. In contrast, our approach achieves effective information hiding and consistency maintenance by decoupling topic shifting and data replacement, along with distillation from the strong model.
Examples of the synthesized query for different methods can be found in Appendix.
% Notably, the Phi3 model outperforms Llama3.2 in the Hint-based method, while Llama3.2 excels in the example-based method.
% However, both models still show higher leakage than our approach. 
% These results demonstrate that our method more effectively controls the flow of sensitive information, mitigating the risk of unintended exposure or leakage.

% \textbf{Generalization of our protection method.}

Our local synthesis model and overall framework are general, maintaining both model-agnostic and dataset-agnostic properties.
Finally, our method demonstrates strong generalization across different datasets and local retrievers. Specifically, the local synthesis model was only trained on the MultiHiertt dataset using context shortened by the local retriever model Phi3. 
% To demonstrate its generalization capability, we directly apply the distilled local synthesizer to the FinQA dataset and evaluate it using the Llama3.2-3B-Instruct model. 
As shown in Table~\ref{table: result_multi}, our approach consistently achieves the highest accuracy and the lowest local data leakage rate, highlighting its effectiveness across unseen datasets (FinQA) and local retriever models (Llama3.2-3B-Instruct).

\subsection{Results with a Referral Module in Model Cascade}
\label{sec:referral}

\begin{figure}
    \centering
    \includegraphics[]{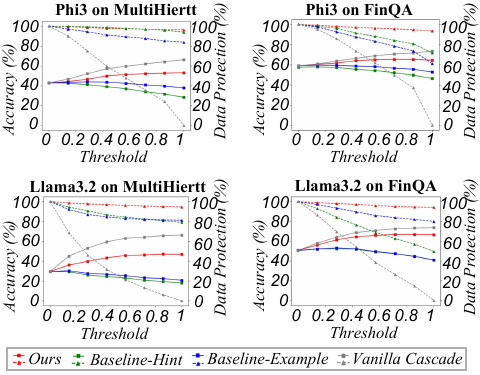}
    % \vspace{-10mm}
    \caption{Model cascade with a decision maker. A higher threshold value results in more instances of seeking remote model collaboration. The solid lines represent accuracy, while the dashed lines indicate local data protection.}
    % \vspace{-5mm}
    \label{fig: cascade}
\end{figure}

In practice, it is unnecessary to forward all local instances to the remote model, as the local model can handle certain simple instances. Building on the vanilla model cascade \cite{yuelarge}, we use the local model's answer consistency rate to determine whether the remote model should be involved. 
% If the consistency rate exceeds a predefined threshold, it indicates that the local model is confident, and the most consistent local answer will be selected as the final answer. 
If the consistency rate is below the threshold, the remote model is activated to assist in the decision-making process. When the threshold is set to 0, all queries are handled locally; when set to 1, all queries involve the remote model.

Fig.~\ref{fig: cascade} illustrates the accuracy and local data protection of various model collaboration methods at different thresholds. The solid lines represent accuracy, while the dashed lines indicate local data protection, with the local data protection rate calculated as 1 minus the leakage rate. The upper-right corner represents a combination of high accuracy and high local data protection. Our results show that while the vanilla model cascade method achieves higher accuracy by directly exposing the full original context, it leads to significant local data leakage. In contrast, baseline privacy-preserving methods improve local data protection but at the cost of dramatically reduced accuracy. Our approach strikes a balance, achieving comparable accuracy while maintaining high local data protection.

\subsection{Ablation Study}
\label{sec:ablation}
To evaluate the effectiveness of the main components of our method, synthesizer distillation and answer reconstruction, we perform an ablation study and present the results in Table~\ref{table: ablation}.

In the "w/o Tool" setting, we replace the tool-based answer construction with local inference using the same synthesized example and remote solution as before ablation. The request synthesis step remains unchanged, but the local model performs inference by combining the synthesized query, remote solution, and original query. This results in a significant accuracy drop, as the local model, with its limited reasoning capability, struggles to comprehend and replicate the example for effective reasoning despite receiving identical remote information.

In the "w/o Distillation" setting, we remove the distillation process and directly prompt the unmodified Llama3.2-3B-Instruct as the local synthesizer using the same data protection instructions. Other components, including tool-based answer reconstruction, remain unchanged.
We observe a significant drop in accuracy, primarily because the local model alone fails to generate logically consistent queries and preserve the original numerical values. 
Firstly, this results in unreliable queries, preventing the remote model from producing a valid problem-solving pattern. Consequently, the tool-based answer reconstruction, which relies heavily on this pattern, fails to generate correct answers. 
Secondly, the unpreserved numbers in the context after topic rewriting cause the local model to switch back to the wrong numbers even if using the correct problem-solving pattern.

In the "w/o Distillation and Tool" setting, we remove both the tool-based answer reconstruction and distillation for the local synthesizer. The local model uses the same synthesized query as in the "w/o Distillation" setting, along with the synthesized example and solution from the remote model, to perform re-inference.
Accuracy improves compared to "w/o Distillation" because, despite the incorrect number in the example, local inference relies more on the model's understanding and reasoning of the original query. However, accuracy remains lower than our method, as incoherent examples and the limited reasoning ability of the local model hinder effective inference.


\begin{table}[]
\centering
\begin{tabular}{ccccc}
\toprule
{}                     & {\textbf{Distil.}} & {\textbf{Tool}} & {\textbf{Acc}}       & {\textbf{Leak.}}  \\ \hline
{Ours}                 & {\ding{51}}             & {\ding{51}}            & {\textbf{90.1}} & {\textbf{5.9}} \\ \hline
{w/o Tool}             & {\ding{51}}             & {\ding{55}}            & {65.0}          & {5.9}          \\ \hline
{w/o Distil.}          & {\ding{55}}             & {\ding{51}}            & {42.2}          & {15.1}          \\ \hline
{w/o Distil. and Tool} & {\ding{55}}             & {\ding{55}}            & {65.7}          & {15.1}          \\ \bottomrule
\end{tabular}
\caption{Ablation study conducted on the FinQA dataset with Llama3.2-3B-Instruct as both the local inference model and local hider model.}
\label{table: ablation}
\end{table}

% \subsection{Our method easily recovers the answer although limited local model ability}
% To evaluate the effectiveness of our data recovery method, we conduct comparison with golden hints or equal examples.

% \textcolor{red}{TODO golden Hint} 
% To study the data recovery ability for hint-based method, we need to avoid the affects of unreliable descriptions. Because the local model may generate a bad description to seek help from the remote model, which will affect the overall reasoning performance. 
% So, given the original question, We prompt the remote LLM to generate high-level hints. We use the golden hint to prompt the local model to generate the final answer. 

% equal example: 
% To study the data recovery ability for example-based method, we need to avoid the affects of unreliable examples. Because the local model may generate a bad example to seek help from the remote model, which will affect the overall reasoning performance.
% We use the same example and solution from our method to prompt the local model to generate the final answer. The difference is that our method uses numeric switch.

\subsection{Sensitivity to Training Data Size}
\label{sec:training_size}

\begin{table}[]
\begin{tabular}{l|l|l}
\hline
\textbf{Training Size} & \textbf{Acc (\%)} & \textbf{Leakage (\%)} \\ \hline
800                    & 74.3                    & 9.2          \\ \hline
2000                   & 73.2                    & 8.2          \\ \hline
5762            & 70.6                    & 5.5          \\ \hline
\end{tabular}
\caption{Sensitivity to training data size on MultiHiertt dataset with Llama3.2-3B-Instruct model.}
\label{table: training_size}
\end{table}

% We varied the size of the training set for topic shifter and showed the results in the following table. We found that as we reduced the training size, the task accuracy remains mostly unchanged, but the leakage consistently increased. Intuitively, the topic shifter still manages to follow the instruction that it should preserve the problem-solving logic, but becomes less good at hiding sensitive information, a task that was likely not particularly optimized for during the model's original training process. This highlights the importance of fine-tuning a local model for the task of transforming the query to protect sensitive information.

We vary the training size for the topic rewriter and report the results in Table~\ref{table: training_size}. As the training size decreases, we observe a slight increase in task accuracy accompanied by a consistent rise in leakage. This is because the topic rewriter sometimes makes only minor changes to the original query when trained on less data. Thus, the accuracy increases at the cost of reduced privacy protection. This aligns with the inherent accuracy–leakage trade-off.
Remarkably, even with only 800 training examples, our method still significantly outperforms existing baselines in both accuracy and privacy, demonstrating the effectiveness of our approach.

\subsection{Error Analysis}
\label{sec:error_analysis}

\begin{table}[]
\begin{tabular}{l|c}
\hline
\multicolumn{1}{c|}{\textbf{Error Type}} & \textbf{Percentage (\%)} \\ \hline
Retrieval Error                        & 20                       \\ \hline
Rewrite Error                         & 16.7                     \\ \hline
Numeric Switch Error                     & 6.7                      \\ \hline
Remote LLM Error                      & 33.3                     \\ \hline
Answer Reconstruction  Error                       & 0                        \\ \hline
Annotation Error/Ambiguous Question   & 23.3                     \\ \hline
\end{tabular}
\caption{Error analysis.}
\label{table: error_analysis}
\end{table}

% We randomly select 30 bad cases from MultiHiertt with the Phi3 model and manually analyze the error types. The error distribution is shown in Table~\ref{table: error_analysis}.

We randomly sample 30 error cases from all incorrect predictions on MultiHiertt with the Phi3 model and manually analyze the error types. The distribution is shown in Table~\ref{table: error_analysis}.

The most common source of error is remote LLM’s error (33.3\%), including evidence grounding, logical reasoning, and Python coding errors. This is the imperfect model’s inherent error, note that the accuracy for remote-only method on MultiHiertt/FinQA dataset is 66.4\% and 73.7\%. 
A significant portion of the errors is attributed to flaws in the dataset itself (23.3\%). Either the annotations were incorrect, or the questions were too ambiguous for a precise answer to be generated.
The retrieval issues account for 20\% of errors. We follow existing methods to shorten the context for long-text reasoning for input length-limited models as the first step in our method. The retrieval is not the main focus or contribution of our paper. It can be replaced with advanced retrieval methods to mitigate retrieval errors. 
The errors for two core designs in our method — topic rewriter (16.7\%) and numeric switch (6.7\%) for answer reconstruction (0\%) are relatively small. Rewriter errors mainly stem from the rewriter model confusing some phrases that look similar or messing up the calculation units. Although imperfect, our design significantly reduces the errors compared to existing baselines by generating logically consistent synthesis data and reusing the Python tool with the data switch.

\section{Conclusion}
% Model cascading utilizes remote large models to address challenges the local model cannot handle, greatly boosting performance over relying solely on the local model. However, this approach transmits raw local data to the remote black-box model, resulting in potential data leakage. Although certain local data protection methods have been successful in tasks such as text classification and writing, they face difficulties in tackling complex logical reasoning problems. 

In this work, we propose a simple yet effective method for LLM numerical reasoning and protect the local data.
% To balance reasoning accuracy and data protection, we design a simple yet effective approach integrating two key components: 
We develop a context-aware synthesis strategy that ensures logical consistency while shifting query domains and a tool-based answer reconstruction approach that reuses remote-generated code snippets with local data. 
Our experimental results confirm the effectiveness of this approach, demonstrating substantial improvements in reasoning accuracy while mitigating data exposure. 
These findings pave the way for further advancements in privacy-preserving LLM collaboration systems, highlighting the potential for secure and efficient deployment in real-world applications.

% In terms of limitations, the method proposed in this paper has been only validated on financial datasets. Experiments on datasets from other domains, such as healthcare, are lacking. This limitation is primarily due to the scarcity of long-text mathematical reasoning datasets in these areas.

% \input{_tex/8_limitations}

\clearpage

\bibliography{refs}

\clearpage
% \input{_tex/ReproducibilityChecklist}

% \clearpage
\appendix
\section{Appendix}
\label{sec:appendix}

\subsection{Experiment Setting}
\label{sec:appendix-dp}
This paper specifies the computing infrastructure used for running experiments (hardware and software), including GPU/CPU models; amount of memory; operating system; names and versions of relevant software libraries and frameworks

\subsection{Why is the DP-based method unsuitable for our task?}
\label{sec:appendix-dp}
The Differential Privacy (DP)-based method \cite{tong2023inferdpt, xie2024differentially} for black-box inference adds noise to generate similar words as replacements for the original words. However, the resulting text is often unreadable and lacks logical coherence. While this approach may perform well on tasks like classification and sentiment analysis that primarily rely on semantic understanding, it is unsuitable for reasoning tasks that require logical consistency. An example generated using InferDPT \cite{tong2023inferdpt} is provided below.

Original text:

\textit{He 's been waiting 19 years for a visa still stuck in a backlog...}

DP-generated text:

\textit{female declared billing 142 Pour must Fantasy Even dear poon am...}

\subsection{Model Cascade Framework}
\label{sec:model-cascade}
The model cascade framework firstly resolves the problem by the local model, if the local response is not reliable, the local problem will be routed to the powerful remote model to resolve. The framework not only reduces the cost of API calls for remote model but also keeps some problems resolved locally which protects the local data from data leakage.

Given the context \( C \) and question \( q \), we prompt the local LLM, denoted as $\mathcal{M_{L}}$ , to perform local inference. To improve robustness, the local model can conduct the inference \( n \) times:
\begin{equation}
    r_i \sim \mathcal{M_L}(C, q)
    \quad \text{for } i \in \{1, \dots, n\} \label{eq:local_model}
\end{equation}
where \( r_i \) represents the solution of the local model at \( i \)-th time.

Then we extract a set of candidate answers \( \mathcal{A} \) from responses \( \{r_{1:n}\} \):
\begin{equation}
    \mathcal{A} = \{ a_1, a_2, \dots, a_n \} \label{eq:candidate_solutions}
\end{equation}

The self-consistency is commonly used to find the final answer and estimate the uncertainty of the answer using voting score.
Given the set of candidate answers \( \mathcal{A} \), we define the consistency score as:
\begin{equation}
S = \max_{a \in \mathcal{A}} \frac{\sum_{i=1}^{n} \mathbb{I}(a_i = a)}{n}
\end{equation}

where \( \mathbb{I}(\cdot) \) is the indicator function that equals 1 if $a_i = a$ and 0 otherwise.

A referral module is then used to determine the reliability of the local answer. we use the voting score of the most common answer as the criterion to decide whether the question needs to be further solved by the remote model \cite{yuelarge}. 
If the consistency score \( S \) is lower than a predefined threshold \( \tau \),
then the remote model $\mathcal{M_R}$ is invoked to assist with reasoning and provide a more reliable answer. However, the direct exposure of local data (context \( C \) and query \( q \)) raises concerns about local data leakage. Observing this data leakage issue, previous work has proposed sharing problem descriptions and semantically similar examples with the remote model. While these methods work well in many cases, we found them ineffective for numerical reasoning tasks. Since the remote query is a fuzzy version of the local one, the returned solution is often not directly applicable or not sufficiently related for the local model to extrapolate, severely degrading the performance of model cascades.

\subsection{Datasets details}
\label{sec:appendix-datasets}

We focus on numerical reasoning over documents for evaluation, so we extract numerical reasoning questions from the original dataset as soon as the answer format is a numerical value.
The FinQA set consists of 1097 samples, while the MultiHierttQA set contains 749 samples. Due to the inconsistent answer formats in the FinQA dataset, such as the inclusion of special symbols and varying decimal places, we execute the given programs in the dataset and retain results with five decimal places.\footnote{\url{https://huggingface.co/datasets/gagan3012/finqa-updated}} 
We convert the table data in the context into textual descriptions using a unified format: ``[row] of [column] is [value].''

\subsection{Training Data Quality Verification for Model Distillation}
\label{sec:appendix-datafilter}
% The MultiHiertt dataset contains a total of 6,360 training examples. Our data filtering process consists of three steps: leakage evaluation, conflict evidence detection, and answer consistency verification.
% In step 1, we randomly select 300 synthesized examples for the leakage check and find that the error rate is 4\%, which is acceptable. In step 3, we manually verify answer consistency and determine that the results are satisfactory. Therefore, we retain all data from steps 1 and 3 to minimize API call costs.
% Step 2 involves conflict evidence detection using a rule-based approach, which results in a 10\% error rate. Based on this, we perform data filtering in step 2. Ultimately, we retain 5,762 examples out of the original 6,360.

The MultiHiertt dataset contains a total of 6,360 training examples. Our data filtering process consists of three steps: leakage evaluation, conflict evidence detection, and answer consistency verification.
In step 1, we randomly select 300 synthesized examples for the leakage check and find that the error rate is 4\%. Step 2 involves conflict evidence detection using a rule-based approach, which results in a 10\% error rate. In step 3, we manually verify answer consistency and determine that the results are satisfactory. Since the numerical values remain unchanged during the topic shift, a logically consistent rewritten question should yield the same answer as the original question. Based on this, we perform data filtering in step 2. Finally, we retain 5,762 examples out of the original 6,360.

\subsection{Numerical Data Transformation}
\label{sec:numeric-data}

For numeric transformation, we employ three strategies: special number handling, offset transformation for year-related values, and order-preserving transformation. 

First, special numbers (e.g., integers such as 28–31, which may indicate the last days of a month, or 1 and 12, which may represent the first and last months of a year) are left unchanged to preserve their semantic meaning.
Second, for integers in the interval [1990, 2030], which are likely to correspond to calendar years, we apply an offset-based transformation. After sorting these values, we randomly assign a base year to the first number and then add relative offsets (e.g., +1, +5) to the subsequent values. For example, the sequence 2003, 2004 (+1), 2008 (+5) might be mapped to 2010, 2011 (+1), and 2015 (+5), respectively. Numbers outside this range are transformed using a sorted random mapping within a target range, excluding the special numbers mentioned earlier.
Third, for all other numbers, we preserve their original order by first sorting them and dividing them into intervals. For each interval, we randomly sample values from the target range and map the original numbers to the corresponding sorted values, maintaining relative ordering.

\subsection{Examples for data leakage}
\label{sec:appendix-leakage examples}
% ----ori_id:  a4301e057a4b49ee9c7ddfcd94250324
We show the synthesized query generated by different methods. The text highlighted in red means leaked data, while the text highlighted in blue means that the data has been protected. 

\textbf{Original query:}

\textit{Context: ...Excluded from the \$33 billion of collateral is \$10 billion of collateral delivered by clients at the initiation of transactions; this collateral secures exposure that could arise in the derivatives portfolio should the MTM of the client’s transactions move in the Firm’s favor.
\\
...Total of Notional amounts 2005 is \$43,593 .
...Total of Derivative receivables MTM 2005 is 50 .
\\
Question: for the derivative contracts , assuming an average contract life of 10 years , what would annual exposure be in us billion at december 31 , 2005 on derivative receivables?}

\textbf{Hint-based method query:}

\textit{This is a question about calculating the \textcolor{red}{annual exposure} of \textcolor{red}{derivative contracts} based on the \textcolor{red}{total notional} amount and the \textcolor{red}{average contract life.}}

\textbf{Example-based method query:}

\textit{Context: ... Excluded from the \textcolor{blue}{\$42} billion of \textcolor{red}{collateral} is \textcolor{blue}{\$12} \textcolor{red}{billion of collateral delivered by clients at the initiation of transactions; this collateral secures exposure that could arise in the derivatives portfolio should the MTM of the client’s transactions move in the Firm’s favor.
... Total of Notional amounts} \textcolor{blue}{2009 is \$57,193} .
\textcolor{red}{Total of Derivative} \textcolor{blue}{payables} \textcolor{red}{MTM} \textcolor{blue}{2009 is 60} .
\\
Question: for the \textcolor{red}{derivative contracts , assuming an average contract life} of \textcolor{blue}{8} years , what would annual \textcolor{red}{exposure} be in us billion at december 31 , \textcolor{blue}{2009} on \textcolor{red}{derivative} \textcolor{blue}{payables}?}

\textbf{Our query}

\textit{Context: ...Excluded from the \textcolor{blue}{46 billion liters of water consumption} is 10 billion \textcolor{blue}{liters} of \textcolor{blue}{water reserved} by clients at the initiation of \textcolor{blue}{usage; this reserved water ensures supply security in case the consumption rate aligns with expectations.}
...Total of \textcolor{blue}{planned water usage 2013 is 35,712 billion liters}.
Total of Actual \textcolor{blue}{water consumption 2013 is 60 billion liters}.
\\
Question: for the \textcolor{blue}{water usage}, assuming an average \textcolor{blue}{usage period} of 10 years, what would annual \textcolor{blue}{water consumption be in billion liters} at December 31, \textcolor{blue}{2013} on actual \textcolor{blue}{water consumption}?}

% 11111
% %ori_id JPM/2018/page_90.pdf-6

% Context A:

% [Sentence 22]: Table (year ended december 31 ( in millions except rates )) shows less : \textcolor{red}{cib markets net interest income} ( c ) of 2018 is 3087

% [Sentence 23]: Table (year ended december 31 ( in millions except rates )) shows less : \textcolor{red}{cib markets net interest income} ( c ) of 2017 is 4630

% [Sentence 24]: Table (year ended december 31 ( in millions except rates )) shows less : \textcolor{red}{cib markets net interest income} ( c ) of 2016 is 6334

% Context B:

% This is a question about calculating the percentage of a specific category (\textcolor{red}{cib markets net interest income}) in relation to another category (managed interest income) for a specific year.

% 22222

% % ---ori_id AMT/2016/page_49.pdf-2
% Context A:

% [Sentence 22]: on february 17 , 2017 , the closing price of our common stock was  108.11 per share as reported on the NYSE .

% [Sentence 23]: as of february 17 , 2017 , we had 427195037 outstanding shares of common stock and 153 registered holders .

% [Sentence 26]: we have two series of preferred stock outstanding , 5.25\% ( 5.25 \% ) mandatory convertible preferred stock... 

% Context B:

% This is a question about determining the market capitalization of a company based on the closing share price and the number of outstanding shares on a specific date.

\subsection{Examples for logic inconsistency}

\textbf{Original query}

\textit{\textcolor{red}{Tier 1} capital of JPMorgan Chase Bank, N.A. Basel III Standardized Transitional Dec 31, 2017 is 184375 .
\\
\textcolor{red}{Total} capital of JPMorgan Chase Bank, N.A. Basel III Standardized Transitional Dec 31, 2017 is 195839 .}

\textbf{Synthesis query with logic error}

\textit{\textcolor{red}{Total} output of Global Manufacturing Division, N.A. Basel III Standardized Transitional Dec 31, 2017 is 184375.
\\
\textcolor{red}{Total} output of Global Manufacturing Division, N.A. Basel III Standardized Transitional Dec 31, 2017 is 195839.}

The synthesis query presents conflicted statements because of logical inconsistency in the rewriting.

\subsection{Prompts and examples for local inference and retrieval}
\label{sec:appendix-local_retriever}

Through local inference with evidence retrieval, we prioritize solving problems locally whenever possible. For cases requiring remote resolution, we compress the long document to alleviate the burden of information hiding for lengthy texts.

\textbf{Evidence localization.} 
The long context makes the local model hard to understand the context and logic thus hinders the generation the generation of similar synthesized problems. So we first shorten the original context via local evidence localization to reduce the burden of synthesizing a similar problem. 

We incorporate evidence localization into the local inference process in the traditional model cascade framework. In this way, the evidence localization will not introduce additional computation overhead. Specifically, we prompt the local model to explicitly display the original sentence from the context as the evidence for each reasoning step. For example, the first evidence for the example in Fig 1 is that ``The aircraft fuel expense in 2018 is 9896.''

The local inference  Equation~\eqref{eq:local_model} will become:

\begin{equation}
    E_i,r_i \sim \mathcal{M_L}(C, q)
    \quad \text{for } i \in \{1, \dots, n\} \label{eq:local_model}
\end{equation}

% \begin{equation}
%     E_i, R_i = \text{LLM}_l(C, Q)
%     \text{ for } i \in \{1, \dots, n\} \label{eq:local_retrieval}
% \end{equation}
where \( E_i \) represents all retrieved sentences at \( i \)-th time.
\begin{equation}
    E_i = \{ s_{i1}, s_{i2}, \dots, s_{ij} \} \label{eq:retrieval}
\end{equation}
We find that the local model sometimes changes the original sentence in the evidence generation because of its hallucination, we just need the model output the sentence id and then use the sentence id to retrieve the original sentences. Table~\ref{table: prompt_local_infer} shows our prompt and an example of the local inference with evidence localization.
Sometimes, the sentence ids may be wrong, so we further use regular expressions to match original sentences containing numerical values in the response as the retrieval results. We remove duplicate retrieved sentences and concat them as the retrieved context. The original long context is shorted as the retrieved context $\mathcal{E} = \bigcup_{i} E_i$.
% \begin{equation}
%     \mathcal{E} = \ concat(set(E_1, E_2, \dots, E_n))
% \label{eq:retrieval}
% \end{equation}

% \begin{equation}
%     \mathcal{E} = \bigcup_{i} E_i
% \label{eq:retrieval}
% \end{equation}

\begin{table*}[htp]
\begin{tabular}{p{1\textwidth}}
% \begin{tabular}{p{\linewidth}}

\hline
\textbf{Prompt}        \\ \hline
You are a helpful assistant. Given the context and the problem, you need to write Python code to solve the problem. Your solution should follow task instructions. \\

Task instruction:\\
Each sentence in the context is numbered starting with [Sentence id]. Your task is to read the given context and write Python code with accompanying comments to solve the question. The Python solution must be enclosed within a code block that starts with the tag ```python and ends with the tag ```.\\

Requirements:\\
1, Break the question down into as many detailed steps as possible.\\
1, In your code comments, specify the type of each step, describe the subtask, and reference the relevant sentence number from the context or provide logical reasons.\\
2, Step Type: Indicate whether the step is a [retrieval step] or a [logical step].\\
3, Retrieval Steps: Indicate all the related sentence number for retrieval with format [Sentence id], [Sentence id]. \\
4, If the retrieved object involves comparison across several candidate sentences, you must NOT directly output the final answer to the retrieval. You need to make sure to list all relevant sentences before outputting the final retrieved objects.\\
5, Logical Steps: Provide the reason or logic behind the decision or action in the code.\\
6, Precision: Retain the full precision of the final result without rounding or truncating decimals.\\
7, Naming Conventions: Python variable names must not start with a number. \\
\hline

\textbf{Example Input}        \\ \hline
Context: \\
{[Sentence 0]}: income was due primarily to the adoption of Statement of Position ...\\
{[Sentence 11]}: Equity securities of For the years ended December 31, 2015 is 17 . \\
...\\

Question: What is the growing rate of Mortgage loans in the year with the most Equity securities? 
\\ \hline

\textbf{Example Output}        \\ \hline
\begin{Verbatim}[commandchars=\\\{\}, codes={\catcode`$=3\catcode`^=7}]
```python
# Step 1, [retrieval step] Retrieve all the related amounts of Equity securities 
\color{red}# evidence from the original context: [Sentence 7], [Sentence 9], [Sentence 11]
equity_securities_2017 = 24  
equity_securities_2016 = 22  
equity_securities_2015 = 17  

# Step 2, [logical step] Determine which year has the most Equity securities amount
year_most_equity = 2017

# Step 3: [retrieval step] Retrieve the mortgage loans amount for the identified year 
2017 and its previous year 2016.
\color{red}# evidence from the original context: [Sentence 13], [Sentence 15]
mortgage_loans_2017 = 124
mortage_loans_2016 = 116

# Step 4: [logical step] Calculate the growth rate of mortgage loans in the year 2017.
# Growth Rate = (Current Year Amount - Previous Year Amount) / Previous Year Amount
growth_rate = (mortgage_loans_2017 - mortage_loans_2016) / mortage_loans_2016
``` 
\end{Verbatim}
\\ \hline

\end{tabular}
\caption{The prompt and an example for local inference and retrieval.}
\label{table: prompt_local_infer}
\end{table*}

% \begin{table*}[htp]
% \begin{tabular}{p{1\textwidth}}
% \hline

% \begin{Verbatim}[commandchars=\\\{\}, codes={\catcode`$=3\catcode`^=7}]
% ```python
% # Step 1, [retrieval step] Retrieve all the related amounts of Equity securities 
% # evidence from the original context: [Sentence 7], [Sentence 9], [Sentence 11]
% equity_securities_2017 = 24  
% equity_securities_2016 = 22  
% equity_securities_2015 = 17  

% # Step 2, [logical step] Determine which year has the most Equity securities amount
% year_most_equity = 2017

% # Step 3: [retrieval step] Retrieve the mortgage loans amount for the identified year 
% 2017 and its previous year 2016.
% \color{red}# evidence from the original context: [Sentence 13], [Sentence 15]
% mortgage_loans_2017 = 124
% mortage_loans_2016 = 116

% # Step 4: [logical step] Calculate the growth rate of mortgage loans in the year 2017.
% # Growth Rate = (Current Year Amount - Previous Year Amount) / Previous Year Amount
% growth_rate = (mortgage_loans_2017 - mortage_loans_2016) / mortage_loans_2016
% \end{Verbatim}

% \\hline

% \end{tabular}
% \caption{The prompt and an example for local inference and retrieval.}
% \label{table: prompt_local_infer}
% \end{table*}

We show the prompt with a golden example as an in-context learning demonstration for local inference and retrieval in Table~\ref{table: prompt_local_infer}.

\subsection{Prompts and examples for topic rewriter}
\label{sec:appendix-topic-shifter}

\begin{table*}[htp]
\begin{tabular}{p{1\textwidth}}
% \begin{tabular}{p{\linewidth}}

\hline
\textbf{Prompt}        \\ \hline
You are a helpful assistant. Rewrite the context and the question according to the following task instruction and requirements.\\
\\
Task instruction:\\
Given the context and question, replace the entities and important nouns in a different topic while keeping the numerical values unchanged.\\
Requirements:\\
1. Change the topic of the context. For example, if the current topic is the water consumption, change it to other topics such as the automobile factory or the medicine.\\
2. Replace important nouns, e.g. securities, issuance, equity compensation plans with general or new-topic-related nouns.\\
3. Maintain the numerical values unchanged.\\
4. Maintain the format and logic unchanged. \\
5. Ensure that every sentence is rewritten. Do not omit or hide any of them. The number of sentences in the output should match the number provided by the user, without omitting anyone.\\
6. Directly output the rewritten context and question within the tag \texttt{<rewritten> and </rewritten>}. Don't include unrelated information in within the tags.\\
\hline

\textbf{Example Input}        \\ \hline
Context:\\
{[Sentence 2]}: Total benefits, claims and expenses increased \$3.9 billion ... and due to increases in the Retail Products segment associated with the growth in the individual annuity and institutional investments businesses.\\
{[Sentence 7]}: Equity securities of For the years ended December 31, 2017 is 24 . \\
{[Sentence 9]}: Equity securities of For the years ended December 31, 2016 is 22 . \\
{[Sentence 13]}: Mortgage loans of For the years ended December 31, 2017 is 124 . \\
{[Sentence 15]}: Mortgage loans of For the years ended December 31, 2016 is 116 . \\
...\\
Question: What is the growing rate of Mortgage loans in the year with the most Equity securities?
\\ \hline

\textbf{Example Output}        \\ \hline
\texttt{<rewritten>}\\
Context:   \\
{[Sentence 2]}: Total expenditures, system failures, and maintenance costs increased by 3.9 billion ... and due to operational expansions in the Electric Vehicle Division to support growing demand in both rural and urban transportation markets.\\
{[Sentence 7]}: Robot units of For the years ended December 31, 2017 is 24 units.  \\
{[Sentence 9]}: Robot units of For the years ended December 31, 2016 is 22 units.  \\
{[Sentence 13]}: Vehicle units of For the years ended December 31, 2017 is 124 units.  \\
{[Sentence 15]}: Vehicle units of For the years ended December 31, 2016 is 116 units.  \\
...\\
Question: What is the growth rate of Vehicle units in the year with the most Robot units?  \\
\texttt{</rewritten>}
\\ \hline

\end{tabular}
\caption{The prompt and an example for the topic rewriter. The original version is mainly about financial and investment context involving retail, mortgage loans. 
The rewritten version shifts the topic to a technological and industry-specific context involving robotics and vehicles.}
\label{table: prompt_topic_rewriter}
\end{table*}

We show the prompt with a golden example as an in-context learning demonstration for the topic shifter in Table~\ref{table: prompt_topic_rewriter}.

\subsection{Why do model collaboration with Hint and Example get worse?}

\begin{figure}
    \centering
    \includegraphics[width=2.5in]{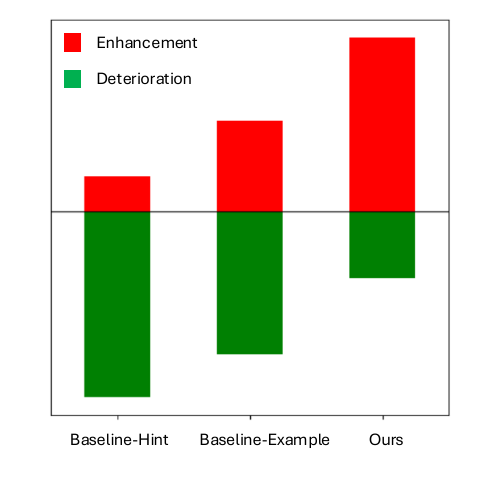}
    % \vspace{-10mm}
    \caption{Enhancements and deteriorations compared to local self-consistency for the Phi3-mini model on the MultiHiertt dataset.  Enhancement refers to cases where local self-consistency was incorrect, but the new method is correct. Deterioration refers to cases where local self-consistency was correct, but the new method is incorrect.}
    % \vspace{-5mm}
    \label{fig: change}
\end{figure}

Our method successfully corrected the cases that the local model initially got wrong, whereas other methods were unable to achieve this.
Fig.~\ref{fig: change} shows enhancements and deteriorations compared to local self-consistency for the Phi3-mini model on the MultiHiertt dataset.  Enhancement refers to cases where local self-consistency was incorrect, but the new method is correct. Deterioration refers to cases where local self-consistency was correct, but the new method is incorrect.
We can observe that the baseline Hint and Example methods caused many questions that could be correctly answered by the local model alone to be answered incorrectly in the collaboration. In contrast, our method successfully addressed a large number of questions that the local model could not solve, thereby improving accuracy.

This is because the unreliable, logic-incoherent query fails to trigger the remote large model for tailored assistance, and it does not effectively invoke the weaker local model's reasoning capabilities.
In contrast, our method generates logically consistent queries to elicit the remote model to produce problem-solving patterns, and it easily recovers answers through data replacement. 

% \subsection{Further explanation for ablation study}

% Originally, the equation was 10 + 1 = 11. The topic-rewriter, whether with or without distillation, expects the values to remain unchanged, so the optimal query after topic rewriting should still be 10 + 1 = 11. However, due to the inaccuracy of the topic-rewriter, the query after topic rewriting becomes incorrect, changing to 10 + 2 = 12. The tool-based answer recovery process involves three steps: switching data to hide the original numbers, getting the remote response for the problem-solving pattern, and switching the data back to the original numbers to be executed for the answer. 

% In the second row, "w/o distill," an ablation study on the local hider is performed, retaining the tool recovery process. The tool recovery restores the value to the state before the topic rewriting, but since the topic-rewritten number is incorrect (10 + 2 = 12), the tool recovery ends up recovering the incorrect value, resulting in 10 + 2 = 12, which has an adverse effect. 

% In the third row, "w/o distill and tool," local inference is used. Despite the incorrect number, local inference relies more on the model's understanding and reasoning of the problem, thus improving performance even with the incorrect input.

% % \subsection{Error source for our method}

% % \textcolor{red}{badcase analysis}

\end{document}